# AI Annotation Orchestration: Evaluating LLM Verifiers to Improve the Quality of LLM Annotations in Learning Analytics

Bakhtawar Ahtisham
Cornell University
Ithaca, USA
ba453@cornell.edu

Kirk Vanacore
Cornell University
Ithaca, USA
kpv27@cornell.edu

Jinsook Lee
Cornell University
Ithaca, USA
jl3369@cornell.edu

Zhuqian Zhou
Cornell University
Ithaca, USA
zz968@cornel.edu

Doug Pietrzak
Freshcognate LLC
Somerville, USA
doug@freshcognate.com

Rene F. Kizilcec
Cornell University
Ithaca, USA
kizilcec@cornell.edu

## Abstract

Large Language Models (LLMs) are increasingly used to annotate learning interactions, yet concerns about reliability limit their utility. We test whether verification-oriented orchestration—prompting models to check their own labels (self-verification) or audit one another (cross-verification)—improves qualitative coding of tutoring discourse. Using transcripts from 30 one-to-one math sessions, we compare three production LLMs (GPT, Claude, Gemini) under three conditions: unverified annotation, self-verification, and cross-verification across all orchestration configurations. Outputs are benchmarked against a blinded, disagreement-focused human adjudication using Cohen's $\kappa$. Overall, orchestration yields a 58% improvement in $\kappa$. Self-verification nearly doubles agreement relative to unverified baselines, with the largest gains for challenging tutor moves. Cross-verification achieves a 37% improvement on average, with pair- and construct-dependent effects: some verifier-annotator pairs exceed self-verification, while others reduce alignment, reflecting differences in verifier strictness. We contribute: (1) a flexible orchestration framework instantiating control, self-, and cross-verification; (2) an empirical comparison across frontier LLMs on authentic tutoring data with blinded human "gold" labels; and (3) a concise notation, `verifier(annotator)` (e.g., `Gemini(GPT)` or `Claude(Claude)`), to standardize reporting and make directional effects explicit for replication. Results position verification as a principled design lever for reliable, scalable LLM-assisted annotation in Learning Analytics.

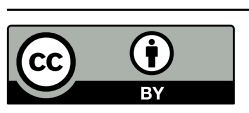

## 1 Introduction

The field of Learning Analytics (LA) has a long tradition of extracting insight from digital traces of learning activity, ranging from adaptive learning systems and learning management platforms to in-person interventions [34]. A central challenge in this work has been the reliable annotation of qualitative data such as tutoring discourse, which encodes pedagogical strategies critical for understanding how students learn. Manual coding remains the standard but is slow, costly, and often yields only moderate inter-rater reliability [17]. Automated approaches using engineered features or supervised Natural Language Processing (NLP) models have improved scalability, but they struggle with the ambiguity and contextual nuance of authentic learning interactions [52].

Large Language Models (LLMs) offer a new pathway for scaling annotation. When guided by codebooks, LLMs can rapidly label interactional moves and sometimes approach human agreement levels [24, 41]. However, early results in LA and Artificial Intelligence in Education (AIED) show mixed performance: reliability varies with prompt design, construct granularity, and evaluation protocol [25]. Moreover, concerns about bias, reproducibility, and evaluator stability remain salient when LLMs are deployed as annotators or judges [23, 50, 53]. These limitations highlight the need for methods that improve the stability and transparency of LLM-assisted qualitative analysis.

One promising direction to overcome these limitations is verification-oriented orchestration, in which models recheck their own outputs (self-verification) or audit one another's labels (cross-verification). Although a single LLM may introduce inconsistency or bias, orchestrating multiple verification steps can reduce idiosyncratic errors by requiring the model to justify or refine its reasoning.

In this way, additional LLMs do not merely compound weaknesses, but can serve as checks and balances that stabilize annotation outcomes. These strategies draw from recent advances in test-time verification methods such as self-consistency and reflective refinement that improve reliability in open-ended tasks [27, 37, 45]. They also build on "LLM-as-a-judge" work, which highlights both the promise and biases of using models as evaluators and motivates the need for careful orchestration [53, 54]. By integrating orchestration strategies with rigorous evaluation, this work bridges LLM-as-judge stability findings [32, 53] with practical pipelines for learning analytics.

To situate our contribution, Table 1 contrasts three common approaches to qualitative coding in education: traditional human coding with two or more coders, single-pass unverified LLM annotation, and our proposed orchestration framework. Human double coding remains the gold standard but is costly and slow, while unverified LLM annotation offers scalability but suffers from instability and prompt sensitivity. Orchestrated verification is fully automated, yet it adapts the logic of human adjudication (self-checks and second-rater cross-checks) into a model-driven pipeline. In doing so, it reframes verification as a principled design parameter for reliable automated annotation.

In this study, we evaluate verification-oriented orchestration for annotating tutoring discourse. Using de-identified transcripts from UPchieve (pseudonymized), a U.S. nonprofit that provides free, on-demand online tutoring, we test Anthropic's Claude 3 Sonnet (Claude), Google's Gemini 2.5 Flash (Gemini), and OpenAI's GPT-5 (GPT). We compare human annotations and unverified LLM annotations from each model against verification-focused orchestration using either the same model (self-verification) or another model (cross-verification). Human baseline annotations in this study were constructed with two human raters using disagreement-focused adjudication (see Methods for details). We structure the contribution of this paper around three guiding research questions:

- **RQ1:** How similar are (unverified) LLM-based annotations of tutoring moves to human reference annotations? To what extent does this depend on the model and the type of pedagogical move?
- **RQ2:** Does verification-oriented orchestration improve annotation reliability? Specifically, how does self-verification compare to cross-verification across models and move types?
- **RQ3:** What systematic patterns emerge in orchestrator verification effectiveness for annotations?

## 2 Related Work

### 2.1 Tutoring Move Annotation

Assigning qualitative labels is a means to deepen our understanding of how tutors and students co-construct learning—turning rich traces of interaction into categories that can test theories about effective help, student agency, and instructional pacing [4, 46, 47]. Because annotations are often categorical and imbalanced (i.e., some categories are rare), reliability is typically reported with chance–corrected indices such as Cohen's $\kappa$ [9] or Krippendorff's $\alpha$ [21]. Interpretive guides provide thresholds for judging whether our operationalizations are stable enough to support claims about tutoring mechanisms rather than annotator artifacts. For example, Landis and Koch [22] describe 0.41–0.60 as moderate and 0.61–0.80 as substantial agreement, whereas McHugh [29] advocates stricter standards (0.80–0.90 as strong). These benchmarks matter not as ends in themselves, but because reliable coding underwrites valid inferences about learning processes [13].

To operationalize tutoring discourse, we draw on categories motivated by learning theory. While there is no single canonical inventory of moves, prior work highlights *scaffolding* [43, 48], *explanations and worked examples* [28, 33], feedback-oriented strategies such as *prompting*, *probing*, and *hinting* [6, 8, 38], and socio–emotional support through *praise or encouragement* [11, 18]. Direct *error correction* also features prominently, though evidence suggests discovery can sometimes be more beneficial [15, 42]. These traditions shape the tutor–moves dictionary used here (Table 2) and foreground why distinctions central to theory (e.g., *Prompting* vs. *Providing Explanation*) can be inherently ambiguous at the utterance level.

LLMs enter this space not as a replacement for theory, but as tools to scale theory-driven inquiry. Under rubric guidance, they have at times approached "substantial" agreement with humans while reducing time-to-analysis [24, 41]. Yet sensitivity to prompt specificity, examples, and construct granularity [25] makes clear that what matters is not raw agreement, but whether we can obtain sufficiently stable labels to test claims about tutoring strategies. Some comparative studies focused on explicit protocols to protect reproducibility and interpretability for theory building. Hybrid human-AI workflows, in which LLMs generate draft categories or exemplars for human adjudication, show promise [2], while comparative evaluations highlight reproducibility and transparency constraints that necessitate explicit evaluation protocols [35, 51]. Taken together, these findings illustrate both the potential and the fragility of LLMs in educational coding, motivating methods that can stabilize reliability without reverting to costly double coding.

### 2.2 Tutoring Analytics as an Annotation Context

One-to-one tutoring compresses instructional moves into short, goal-directed exchanges wherein meaning often lies in intent, timing, and contingency rather than surface form. Labels such as *Prompting*, *Revoicing*, and *Probing Student Thinking* therefore index hypothesized mechanisms (i.e., activation of prior knowledge, uptake, formative assessment) more than mere lexical patterns. Compared to whole-class lessons, one-on-one sessions feature rapid adjacency pairs, elliptical utterances, and high local ambiguity [7, 16], making them a stringent test of whether our coding schemes actually capture the constructs we theorize. Platforms that provide tutoring at scale can offer ecologically valid data streams for probing how these moves function in authentic settings.

Methodological advances in automated discourse analytics should be read through this lens of construct discovery. Early Automatic Speech Recognition (ASR) pipelines [10] and later transfer learning architectures (e.g., BERT) improved recognition over n-gram baselines yet struggled on rare but theoretically critical moves [20]. Curated corpora such as *TalkMoves* enabled comparative modeling anchored in Accountable Talk theory [39], and local-context transformers improved classification [40].

**Table 1: Comparison of three annotation approaches in educational discourse analysis and their expected strengths, weaknesses, and reliability. The verification-focused orchestration approach advanced in this paper is contrasted with traditional human and unverified LLM annotation.**

| | Human Annotation | Unverified LLM Annotation | Verified LLM Annotation (this study) |
|---|---|---|---|
| **Process** | Two or more independent human raters apply rubric; disagreements adjudicated | One model applies rubric once; output used directly | Model(s) apply rubric, then outputs are verified through self- or cross-checks |
| **Expected Strengths** | Gold-standard validity; nuanced interpretation | Scalable, low cost, rapid annotation | Improves stability and reliability; leverages complementary model biases |
| **Expected Weaknesses** | Time- and labor-intensive; limited scalability | Unstable; sensitive to prompt design and construct ambiguity | Added computational overhead; cross-verification benefits are pair- and construct-dependent |
| **Expected Reliability** | Moderate–high, depending on training and construct ambiguity | Variable; often below human agreement | Consistently higher than single-pass; self-verification yields robust gains; cross-verification yields selective improvements |

Recent work has shifted from classrooms to tutoring and from supervised models to LLMs. Studies show that prompting, embedding-based classifiers, and fine-tuning can classify tutor talk moves at scale, though with sensitivity to move type and prompt design [5, 30]. New corpora reveal systematic differences between tutoring and classroom settings (e.g., higher rates of information requests, lower revoicing), challenging model transfer [5]. Complementary work adapts classroom frameworks for 1:1 settings and demonstrates feasibility with transformer baselines [1]. This line of research continues a long tradition in education research of analyzing tutor discourse to understand and support learning [49].

These dynamics motivate our investigation into orchestration for annotation: in tutoring, where turns are short and intent is implicit, unverified LLM annotations can be especially vulnerable to near-miss confusions. Verification-oriented orchestration, with self-checks that force rubric-referenced justification and cross-checks that leverage complementary model biases, offers a pragmatic path to stabilize labels without reverting to multiple human coders.

## 2.3 LLM-as-a-Judge, Verification, and Orchestration

LLMs have been investigated as evaluators, a paradigm often referred to as "LLM-as-a-Judge" [53, 54]. This work extends this construct-centered perspective to evaluation: models can apply rubrics and compare responses at scale, but biases (e.g., verbosity, position effects), scale drift, and run-to-run variance threaten the validity of inferences about underlying abilities or strategies [26, 50, 53, 54]. Mitigation approaches such as rubric guidance, pairwise comparisons, and multi-pass checks are valuable insofar as they preserve the signal we seek to capture: principled differences among strategies, moves, or learning processes. Related multi-agent approaches (debate, role specialization, collaborative orchestration, etc.) similarly aim to increase reasoning reliability so theoretically precise contrasts are not washed out by evaluator noise [12, 55].

Within this study, we treat verification as a method for construct stabilization rather than a benchmark exercise. *Self-verification* prompts a model to re-derive or critique its own labels (e.g., self-consistency, reflective refinement), reducing variance and clarifying the warrant for a move assignment [27, 32, 37, 45]. *Cross-verification* treats a second model as an independent auditor, analogous to a second human coder; it can expose systematic errors and calibrate strictness, though benefits depend on construct and pairing [26, 53, 54]. We distinguish between *construct stabilization* and *inter-rater reliability (IRR)* in interpreting these effects. Verification operates as a *process-level* mechanism for construct stabilization, reducing random variance and rubric drift during annotation by requiring explicit, rubric-grounded reconsideration. Cohen's $\kappa$, by contrast, is an *outcome-level* statistic that quantifies agreement after the annotation process is complete. In this framing, verification explains *how* agreement is produced, while $\kappa$ measures *how much* agreement is achieved.

We define *orchestration* as the structured coordination of such verification passes under a shared, theory-grounded rubric. Our contribution is to operationalize orchestration for fine-grained tutoring moves and to quantify how different verification choices affect the stability of the various constructs. In other words, we ask whether orchestration helps us see the tutoring phenomena more clearly, not merely whether it raises an agreement coefficient.

## 2.4 Current Study and Novelty

We extend prior work in two ways. First, we evaluate orchestration on authentic 1:1 tutoring discourse rather than curated exam responses or essays, targeting fine-grained pedagogical moves. Second, we analyze reliability *by construct and by verifier-annotator pair*, showing that self-verification reliably raises agreement, especially for intent-sensitive categories. While cross–verification can outperform self-verification, it also systematically depresses reliability for certain constructs. This construct-sensitive view reframes orchestration as a design parameter rather than a monolithic choice, aligning with calls for evaluator stability and transparent protocols in LLM-as-a-judge research [53].

# 3 Methods

## 3.1 Tutoring Data and Context

This study analyzes data from a U.S.-based online tutoring service – UPchieve – that connects volunteer tutors with students from Title I schools. Their tutoring operates on an on-demand model: students can request help from a live tutor whenever they encounter difficulty, rather than attending pre-scheduled sessions. Tutoring takes place entirely through a chat system equipped with a shared digital whiteboard. Our analysis focuses on annotating 1,881 utterances from de-identified transcripts of 30 randomly-sampled tutoring sessions. While the platform supports tutoring across more than fifteen subject areas, the present research centers specifically on mathematics sessions at the secondary school level.

## 3.2 Codebook Development

To evaluate how well LLMs can annotate tutoring discourse, we first developed a codebook of tutor moves through an inductive-deductive process aligned with best practices in qualitative coding [3, 4, 46]. First, an inductive review of one-to-one math interactions identified recurring strategies (e.g., prompting, probing, correcting). Categories were defined with explicit inclusion/exclusion criteria and near-miss examples to clarify boundaries. We then aligned these categories with established theory, including Vygotskian scaffolding [43, 48], instructional principles [28, 33], formative feedback [6, 8, 38], socio-emotional support [11], and the feedback and tutoring literature [15, 42]. The codebook was iteratively refined and then used to label each Tutor utterance with a single best-fit move. The final codebook with definitions is shown in Table 2.

## 3.3 Ground Truth Construction

To establish ground truth, we followed a human-AI collaborative annotation procedure adapted from [44] and [24]. Initially, Author 1 manually coded 30 transcripts using the finalized codebook. The same set of transcripts was then processed by *Google's Gemini 2.5 Flash* model using a structured prompt that contained each code's definition and example instances. We compared the two sets of annotations and isolated all instances of disagreement. Author 1 and Gemini disagreed on 501 utterances; 26.63% of the entire sample.

To resolve these disagreements, we produced a new dataset labeling the two coders as "Rater 1" and "Rater 2," randomizing which label corresponded to the human or AI at the utterance level. An external reviewer with experience in human tutoring implementation and research conducted a blinded review (i.e., unaware of the source of the label) of these cases and indicated agreement with one of the two raters. The resulting judgments were nearly balanced, with the external reviewer aligning with Gemini on 53.23% of disagreements. Ground-truth labels were then derived from all instances of agreement between Author 1 and Gemini, as well as from those adjudicated by Author 3. This procedure employs a simpler approach, building upon [44], which demonstrates that human–AI verification pipelines can yield reliable gold-standard data and aligns with emerging research that AI's attentional capacity and lack of fatigue can be leveraged in attaining ground truth, so long as the human remains in the loop [31].

## 3.4 LLM Annotation and Orchestration Setup

We then used three LLMs (**GPT**, **Claude**, and **Gemini**) to annotate the data to compare three conditions to the human annotation baseline:

(1) **Unverified LLM Annotation.** The model assigns one label from Table 2 to each tutor utterance. A brief justification is logged for transparency, not for scoring.

(2) **Self-verification Orchestration.** Building on unverified LLM annotation, the same model is used to review its initial label against the rubric, explicitly checking definitions and examples. It then either retains or revises the label. The final output is the post-verification decision.

(3) **Cross-verification Orchestration.** Building on unverified LLM annotation, a different model (the verifier model) verifies the annotator's label using the rubric, the utterance, and the annotator's rationale. The verifier outputs retain or revise, producing the final label. We tested all six cross-verification pairs of verifier(annotator): GPT(Claude), GPT(Gemini), Claude(GPT), Claude(Gemini), Gemini(GPT), and Gemini(Claude).

Throughout this paper, we refer to the verifier as the LLM responsible for checking the annotations produced by another model. For clarity, we denote this relationship as verifier(annotator). For example, if the notation is written as GPT(Gemini), this indicates that GPT serves as the verifier, evaluating the annotations generated by Gemini. This notation is model-agnostic and can be used in future work to represent different orchestration strategies and combinations, including orchestration over multiple sources of annotation from AI models and/or humans.

All conditions used the same rubric-grounded prompt, which contained category names, operational definitions (Table 2), and concise in-context examples. Transcripts were chunked at discourse-coherent boundaries (median ≈ 80 turns) with 1-2 overlapping turns. For each tutor utterance, the model saw: (i) the focal utterance, (ii) the preceding student turn, and (iii) the tutor's prior turn (when available). For verification conditions, we log (a) the annotator's initial label, (b) the verification decision, (c) the final label, and (d) a minimal rubric-referenced justification.

## 3.5 Evaluation of LLM Orchestrated Annotations

We evaluate the reliability of each orchestration configuration by computing Cohen's $\kappa$ for each tutor-move category under every orchestration configuration and comparing against the ground truth. This measure corrects for chance agreement and is standard in qualitative coding research, allowing direct comparison with prior educational discourse studies [22]. To capture construct-level sensitivity, we report $\kappa$ values per tutor move rather than collapsing into a single aggregate, since overall means can mask performance variation across instructional moves.

To isolate the effect of verification on annotation reliability, we compute per-category improvements relative to the same unverified model's baseline, defined as:

$$\Delta\kappa_{verifier(annotator)} = \kappa_{verifier(annotator)} - \kappa_{annotator} \quad (1)$$

Where $\kappa_{verifier(annotator)}$ represents reliability for orchestration configuration $verifier(annotator)$ and $\kappa_{annotator}$ represents the baseline reliability for a given LLM $annotator$. Thus, positive $\Delta\kappa^{c}$ values indicate increased agreement with human coders when an orchestration step is added to the annotation pipeline.

**Table 2: Tutor-moves codebook used in this study. Abbreviated definitions shown; full rubric with examples appears in the supplementary materials.**

| Tutor Move | Operational definition |
|---|---|
| Emotional Support | Affirming affect, normalizing struggle, conveying empathy or encouragement that supports motivation/engagement (e.g., reassurance, acknowledgement of effort). |
| Giving Praise | Positive evaluative feedback on performance or effort (e.g., "Great job," "Nice approach"), typically tied to the student's recent action. |
| Error Correction | Directly identifies an error and supplies the correct form/answer or explicitly negates an incorrect claim. |
| Providing Explanation | Explains a concept, step, or rationale; may include brief worked-example style demonstrations or step-by-step reasoning. |
| Exemplifying | Introduces a concrete example or counterexample to illustrate a concept or procedure; focuses on the illustrative instance. |
| Giving Hint | A minimal cue or nudge (keyword, next-step pointer) intended to move the solution forward without full rationale. |
| Prompting | An open or leading question that advances the problem-solving process (e.g., "What's your next step?" "How could you isolate $x$?"). |
| Probing Student Thinking | Asks the student to articulate reasoning, justify steps, or compare alternatives (e.g., "Why does that rule apply here?"). |
| Revoicing | Restates or summarizes the student's idea to check understanding or highlight key elements of the contribution. |
| Scaffolding | Provides structured support that breaks the task into subgoals, offers ordered guidance, or fades support across turns. |
| Using Visual Cues | Directs attention to or introduces diagrams/notations/highlighting or other layout cues to support understanding. |

# 4 Results

## 4.1 Baseline Reliability: Unverified LLMs Exhibit Low, Uneven Agreement

We first examine how unverified LLM annotations align with the ground truth across the eleven tutor–move categories. As shown in Figure 1, baseline reliability (*Cohen's* $\kappa$) is low overall and uneven across both categories and models, reflecting the difficulty of annotating intent-sensitive pedagogical moves from short tutoring turns. For all three models (Claude, Gemini, and GPT) agreement rarely exceeds moderate levels, with several categories clustering near chance. This finding reinforces the challenge of applying rubric-based qualitative coding to authentic one-to-one tutoring dialogues without contextual reasoning or iterative checks.

Tutor Moves such as *Prompting*, *Revoicing*, and *Probing Student Thinking* exhibit the weakest alignment, often falling below $\kappa = 0.20$, indicating slight reliability. These moves depend heavily on inferred instructional intent rather than explicit lexical markers and are highly sensitive to discourse context. For instance, a prompt like "What next?" may be labeled as *Prompting*, *Scaffolding*, or even *Probing* depending on preceding turns, subtleties that LLMs often overlook when reasoning in isolation. Misclassifications also occur between semantically adjacent categories such as *Providing Explanation* and *Exemplifying*, where both contain explanatory content but differ in emphasis and purpose.

In contrast, categories with more distinct surface cues, such as *Giving Praise* ("Great job!") and *Emotional Support* ("Don't worry, this is tricky for everyone"), achieve higher agreement ($\kappa > 0.60$), suggesting that lexical regularity and sentiment cues facilitate consistent classification. Mid-range categories such as *Error Correction* and *Providing Explanation* yield low to moderate reliability (0.30–0.50), reflecting clearer structural features (e.g., negation or declarative forms) yet lingering ambiguities in intent.

No model consistently outperforms the others. Claude demonstrates slightly higher reliability for socio-emotional moves, possibly due to stronger sentiment understanding, whereas Gemini shows modest advantages on reasoning-oriented categories such as *Probing Student Thinking*. GPT aligns better on procedural guidance moves like *Providing Explanation* and *Error Correction*. Overall, these results establish a low-performing baseline, motivating the need for verification-oriented orchestration to stabilize and enhance LLM-based annotation.

## 4.2 Verification Improves Reliability but Effects Vary by Construct and Model Pairing

We next assess whether verification-oriented orchestration improves reliability over the baseline. Figure 2 plots the change in $\kappa$ for each model under self- and cross-verification relative to its unverified baseline. Across models, self-verification produces consistent and substantial gains, with median improvements of $\Delta\kappa \approx 0.20$–$0.30$. The act of revisiting one's own prediction under rubric constraints appears to foster more deliberate reasoning and reduce mislabeling. Gains are particularly pronounced for intent-heavy categories, such as *Prompting*, *Probing Student Thinking*, and *Revoicing*, which previously suffered from ambiguity. Self-verification forces explicit justification against the rubric, encouraging consideration of subtle contextual cues and yielding more stable alignment with the ground truth.

For surface-cue categories such as *Giving Praise* and *Emotional Support*, the ceiling effect is evident: the already high baseline reliability leaves less room for improvement, and post-verification gains are modest. This pattern suggests that verification primarily benefits cognitively demanding categories requiring deeper interpretation rather than simple lexical recognition.

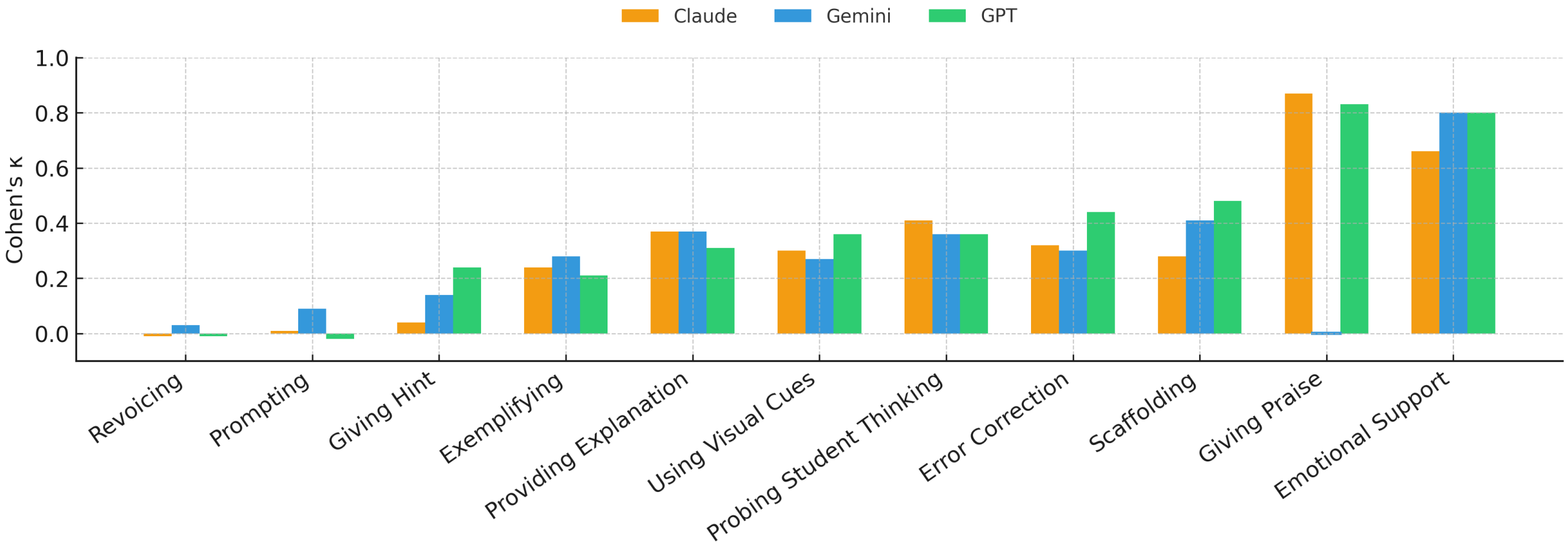

**Figure 1: Reliability of Unverified LLM Annotations by Tutor–Move Category. Bars show Cohen's $\kappa$ with ground truth for Claude, Gemini, and GPT.**

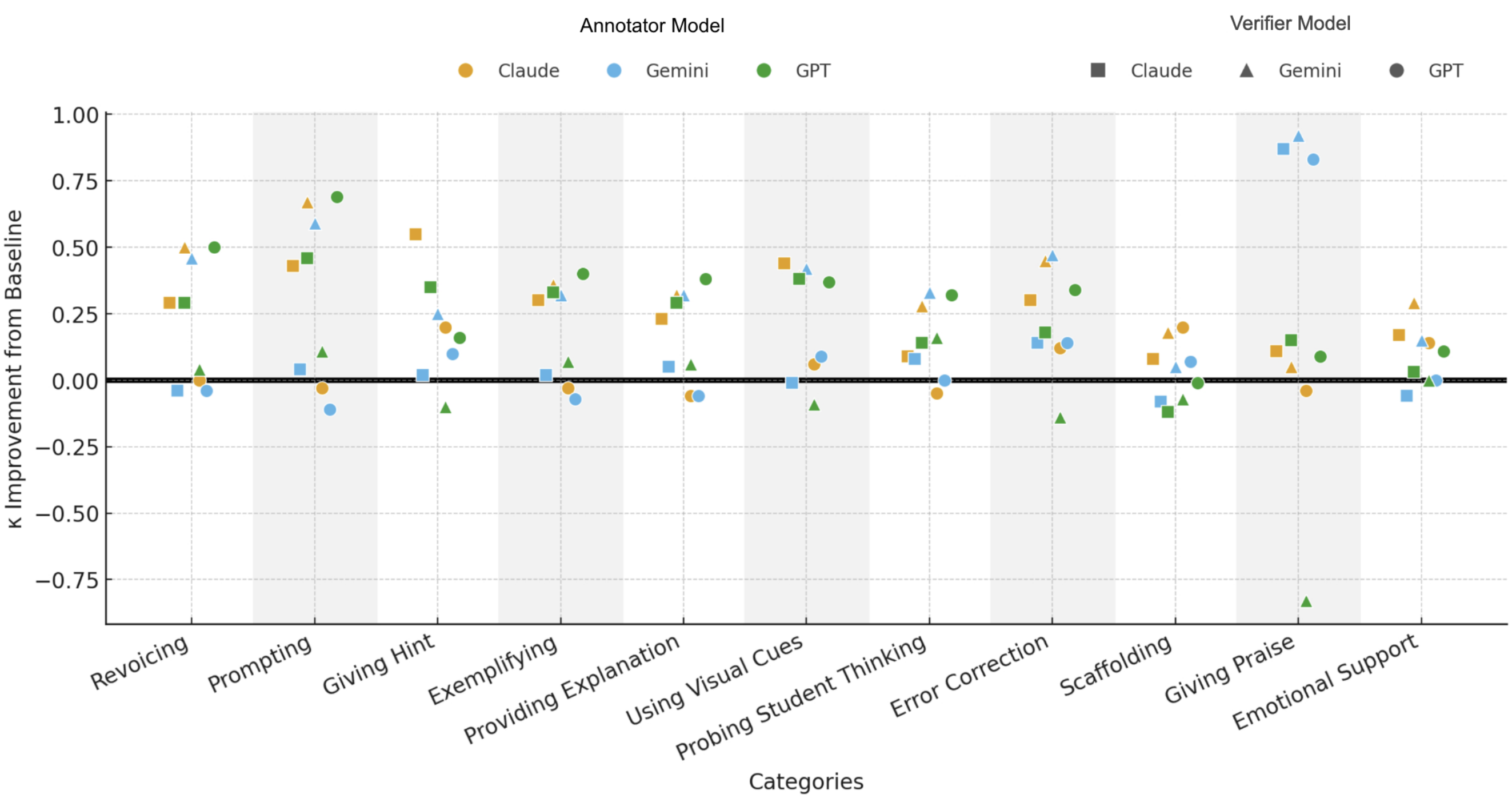

**Figure 2: Improvement in $\kappa$ from the unverified baseline LLM annotation of the corresponding annotator model (e.g., the $\kappa$ improvement for GPT(Gemini) here is relative to Gemini). Each point represents a verifier(annotator) combination; markers above zero indicate improved agreement, while negative values reflect declines.**

By contrast, cross-verification exhibits heterogeneous and directionally asymmetric effects. Some pairings yield additional improvements beyond self-verification, while others reduce reliability. For example, Gemini acting as verifier for Claude often boosts agreement, especially on reasoning-oriented categories like *Providing Explanation* and *Probing*, potentially due to Gemini's greater contextual sensitivity. Conversely, Claude verifying GPT tends to depress scores on several constructs, indicating overcorrection or stricter thresholding. Similarly, GPT checking Gemini occasionally introduces noise when interpretive biases misalign. The dispersion around zero in Figure 2 highlights this variability, suggesting that cross-verification's benefits depend not only on the annotator's initial output but also on the verifier's calibration and rubric interpretation.

## 4.3 Comparative Patterns Across Orchestration Strategies

To better understand orchestration dynamics, Figure 3 presents per-model reliability under all verification strategies. Three broad patterns emerge: First, verification-oriented orchestration substantially increases agreement relative to baseline. Across all categories and models, the average $\kappa$ rose from 0.32 in the unverified LLM annotation baseline to 0.64 under self-verification and 0.44 under cross-verification, yielding an overall verified mean of 0.51. This reflects a roughly 58% overall improvement, with self-verification nearly doubling reliability and cross-verification providing more moderate, pair-dependent gains (+37%).

Self-verification consistently lifts $\kappa$ across nearly all categories and models, confirming it as a stable, low-risk enhancement. All three models show a mean $\kappa$ above 0.60 post-verification, with Gemini reaching 0.77 on average and GPT 0.73, marking a clear shift from low baseline alignment. The reflective check appears particularly effective for intent-driven moves, suggesting that deliberate rubric reapplication reduces impulsive misclassifications.

Second, cross-verification effects are pair-dependent and order-sensitive. Reliability gains vary across model combinations and depend on which model serves as the verifier. Certain pairings, such as `Claude(GPT)` or `Gemini(Claude)`, consistently improve agreement on reasoning-oriented categories (e.g., *Prompting*, *Probing Student Thinking*), suggesting that some verifiers are better tuned to detect and correct these specific misclassifications. In contrast, reversing the order (e.g., `Claude(Gemini)` or `GPT(Gemini)`) often yields lower alignment, indicating that verification direction influences outcomes.

Third, even with verification, construct-level differences persist. Moves such as *Revoicing* and *Prompting* achieve only slight to moderate reliability, highlighting intrinsic ambiguity in detecting instructional intent from text alone. Verification narrows but does not eliminate these gaps, suggesting the need for additional contextual or multimodal signals for full parity with human coders.

## 5 Discussion

Our findings highlight verification-oriented orchestration as a promising method for improving the reliability of LLM-based qualitative coding in LA. Across three current state-of-the-art models, unverified LLM annotations achieved only modest alignment with the ground truth, particularly on intent-sensitive discourse moves such as *Prompting*, *Revoicing*, and *Probing Student Thinking*. These results highlight longstanding challenges in automating educational discourse analysis, where categories rely on inferred pedagogical intent rather than lexical form [7, 15]. Similar to prior studies documenting low inter-annotator agreement for complex instructional constructs [10, 39], our results illustrate that raw LLM judgments are often insufficient for dependable analytics without orchestration.

Introducing verification substantially improved reliability, especially through self-verification, which consistently raised Cohen's $\kappa$ across all models and categories. This pattern supports prior findings that reflective prompting and rubric reapplication enhance model consistency by forcing deliberate reconsideration of borderline cases [27, 37, 45]. The gains were most pronounced in categories with low baseline reliability, suggesting that self-verification mitigates error that may be attributed to hallucinations by promoting alignment with operational definitions. In effect, self-verification functions analogously to second-pass coding that helps AI annotations better align with human-driven ground truth. This offers a scalable and low-cost method to approximate adjudication.

Cross-verification, while sometimes yielding further improvements, depended on model pairing and category. Certain configurations, such as `Gemini(Claude)`, elevated reliability on reasoning-oriented moves, whereas others, like `Claude(GPT)`, occasionally reduced alignment. These directional differences align with prior LLM-as-judge research showing sensitivity to evaluator calibration and positional effects [53, 54]. Our results extend this literature by demonstrating that cross-verification must be treated as a parameterized design choice rather than a universal enhancement. The interplay between base and verifier models shapes whether orchestration amplifies complementary strengths or compounds systematic errors.

Despite these improvements, construct-level reliability followed a stable hierarchy across orchestration strategies: socio-emotional moves (*Giving Praise*, *Emotional Support*) remained easiest to detect, while cognitively complex moves (*Prompting*, *Revoicing*) remained hardest. This persistence suggests that model performance is constrained less by architecture than by intrinsic construct ambiguity, echoing observations from classroom discourse analytics where move boundaries overlap conceptually [19, 40]. Further progress may therefore depend on richer exemplars, multi-turn context windows, or multimodal signals that better capture pedagogical intent.

From a systems perspective, orchestration reframes annotation not as a static output but as an iterative reasoning process. By treating verification direction (`verifier(annotator)`) as a controllable parameter, developers can tune pipelines for different tradeoffs—stability, cost, or interpretability—mirroring the tunable workflow envisioned in multi-agent orchestration frameworks [12, 55]. Compared to human-in-the-loop and active learning methods [14], which reduce but do not eliminate human labor, verification-based orchestration offers a fully automated yet auditable alternative: every judgment is logged and justified. This transparency aligns with emerging norms in LLM evaluation calling for rubric-grounded, explainable, and multi-pass judging [54].

These findings carry practical implications for scaling trustworthy educational analytics. First, self-verification may be adopted as a default mechanism to stabilize LLM coders in high-volume pipelines, particularly for intent-laden constructs. Second, cross-verification should be deployed selectively, guided by empirical pairing performance and construct-level diagnostics. Third, human adjudication can be reserved for persistent disagreements or low-agreement categories, thereby concentrating expert effort where it adds maximal value. This hybrid reliability policy aligns with recommendations from prior learning analytics research, which advocates for strategic automation in annotation-intensive tasks [36].

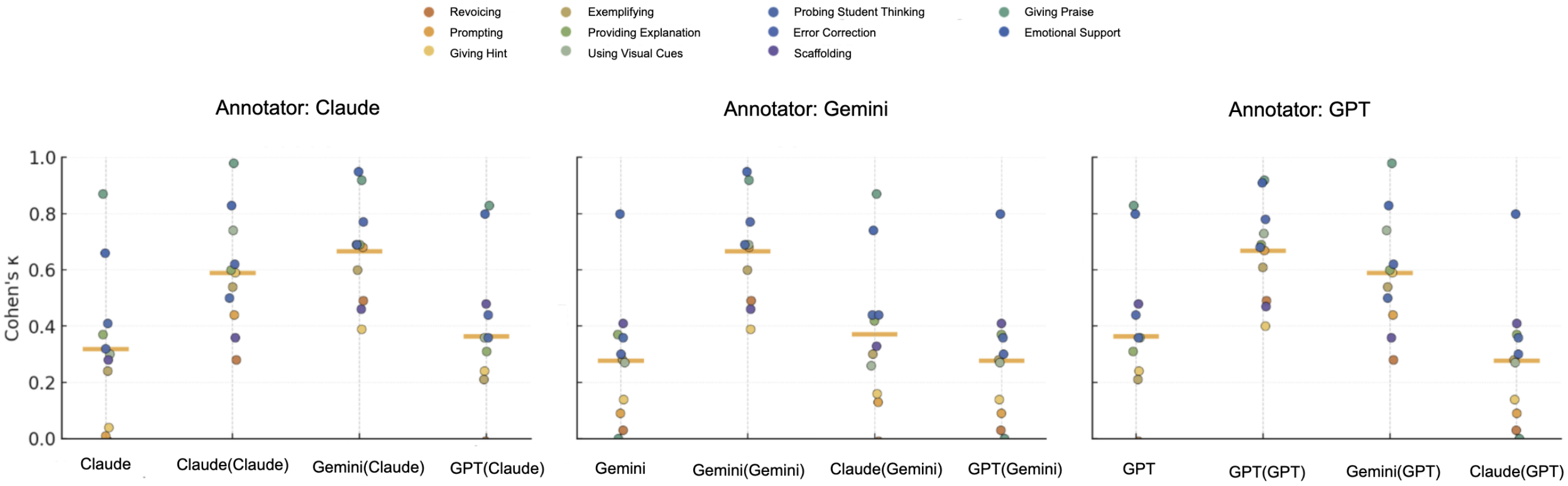

**Figure 3: Per–model reliability under different verification strategies. Each facet fixes the base annotator; points show Cohen's $\kappa$ across categories for baseline, self–verification, and cross–verification directions. Yellow lines denote means.**

Ultimately, verification-oriented orchestration moves the field toward a more principled synthesis of scale and rigor. By coupling reflective reasoning with systematic cross-model calibration, it advances a transparent, evidence-based framework for using LLMs as educational coders—one capable of meeting the dual demands of efficiency and accountability in large-scale tutoring analytics.

## 6 Limitations

This study evaluates verification-oriented orchestration on de-identified, text-chat transcripts from 30 one-to-one math tutoring sessions on UPchieve. While the corpus spans a range of problem types and tutoring styles, its scope remains modest and domain-specific, limiting generalizability across subjects, grade levels, modalities (e.g., audio/video), codebooks, and annotation tasks. We therefore interpret our estimates as evidence that orchestration methods hold promise for scaled annotation in LA; however, more work is needed to understand how these methods replicate and what orchestration configurations function best in which contexts.

Our ground truth adjudicated reference concentrates expert effort by resolving disagreements between an initial human coder and an LLM through a second human rater. Although this method has emerging research that supports it, it could have possibly produced bias towards the LLM annotations. However, since we used Gemini as an annotator when establishing the ground truth – checked by a human – and it did not substantially outperform the other models as an annotator, we do not have evidence that such bias was introduced.

Findings also reflect specific model snapshots and a single rubric prompt family with concise, in-context examples. Provider updates, longer context windows, richer rationales, or alternative prompting strategies may shift both baselines and verification returns. We did not exhaust the design space (e.g., chain-of-thought variants, few-shot scaling, adversarial near-miss exemplars), so orchestration gains should be read as comparative within this prompt regime, not as absolute maxima. In addition, our labels are single-best-fit per tutor utterance; many turns plausibly host co-occurring moves. Multi-label schemes and sequence-aware scoring may alter the observed difficulty gradient, particularly for intent-laden constructs such as *Prompting* and *Revoicing* [40]. Finally, we did not optimize for compute cost or latency. Some cross-verification chains that lift reliability may be impractical in time-sensitive pipelines; deployment will require cost-aware policies that adapt verification depth to uncertainty, construct, and service constraints.

Replication across educational contexts, student and tutor populations, and instructional modalities is a key next step, as is testing whether directional asymmetries in cross-verification generalize to more LLMs. Extending orchestration to speech/video, larger context windows, and richer exemplars, and integrating adaptive policies that select `verifier(annotator)` dynamically based on confidence and construct, are promising directions for robust, equitable learning analytics.

## 7 Conclusion

Orchestrated verification reframes LLM-based qualitative coding as a controllable design space rather than a one-shot prediction task. In authentic one-to-one tutoring discourse, single-pass annotations yield low and uneven agreement with the ground truth reference, especially for intent-sensitive moves. Requiring models to reapply the rubric and justify decisions—self-verification—consistently improves reliability, with the largest gains on the very constructs that are hardest to code. Cross-verification can further raise the ceiling but is directional and construct dependent, highlighting verification *direction* (`verifier(annotator)`) as an explicit design parameter rather than a monolithic choice. Taken together, these results connect LLM-as-a-judge stability concerns to a practical, auditable pipeline for tutoring analytics: enable self-verification by default to harvest robust gains, deploy selected cross-verification pairings where empirical improvements warrant the effort, and reserve human adjudication for persistent edge cases. Beyond tutoring, the same principles (reflection, calibration, and construct sensitivity) provide a template for scalable, transparent annotation in education research and related domains where dependable qualitative coding is essential.

# A Digital Appendix:

https://docs.google.com/document/d/e/2PACX-1vTrpDJe67l-lkPZ-qu_zGNFG8XXfe_BF2Dt98sp4HRCx_5x-bO7NMDbVTVDztJt7uAIFR_Q5uoCEad0/pub